%% file: main.tex
\documentclass[10pt,twocolumn,letterpaper]{article}

\usepackage[pagenumbers]{cvpr} 

\input{preamble}

\definecolor{cvprblue}{rgb}{0.21,0.49,0.74}
\usepackage[pagebackref,breaklinks,colorlinks,citecolor=cvprblue]{hyperref}

\def\paperID{8598} 
\def\confName{CVPR}
\def\confYear{2024}

\title{HEViTPose: High-Efficiency Vision Transformer for Human Pose Estimation}

\author{Chengpeng Wu, Guangxing Tan\thanks{Corresponding author.}, Chunyu Li \\
Liuzhou Key Laboratory of Intelligent Sensing and Control\\
College of Automation, Guangxi University of Science and Technology, Liuzhou, China\\
{\tt\small 221068325@stdmail.gxust.edu.cn, 100000321@gxust.edu.cn,221068310@stdmail.gxust.edu.cn}
}

\begin{document}
\maketitle
\input{sec/0_abstract}    
\input{sec/1_intro}

{
    \small
    \bibliographystyle{ieeenat_fullname}
    \bibliography{main}
}


\end{document}

%% file: preamble.tex
\usepackage[dvipsnames]{xcolor}

\usepackage{makecell}
\usepackage{pifont}
\usepackage{indentfirst}
\usepackage{amssymb}
\usepackage{multirow}

%% file: sec/0_abstract.tex
\begin{abstract}

Human pose estimation in complicated situations has always been a challenging task. 
Many Transformer-based pose networks have been proposed recently, achieving encouraging progress in improving performance. However, the remarkable performance of pose networks is always accompanied by heavy computation costs and large network scale.
In order to deal with this problem, this paper proposes a High-Efficiency Vision Transformer for Human Pose Estimation (HEViTPose).
In HEViTPose, a Cascaded Group Spatial Reduction Multi-Head Attention Module (CGSR-MHA) is proposed, which reduces the computational cost through feature grouping and spatial degradation mechanisms, while preserving feature diversity through multiple low-dimensional attention heads.
Moreover, a concept of Patch Embedded Overlap Width (PEOW) is defined to help understand the relationship between the amount of overlap and local continuity. By optimising PEOW, our model gains improvements in performance, parameters and GFLOPs.

Comprehensive experiments on two benchmark datasets (MPII and COCO) demonstrate that the small and large HEViTPose models are on par with state-of-the-art models while being more lightweight.
Specifically, HEViTPose-B achieves 90.7 PCK@0.5 on the MPII test set and 72.6 AP on the COCO test-dev2017 set. 
Compared with HRNet-W32 and Swin-S, our HEViTPose-B significantly reducing Params (\textcolor{red}{$\downarrow$62.1\%};\textcolor{red}{$\downarrow$80.4\%}) and GFLOPs (\textcolor{red}{$\downarrow$43.4\%};\textcolor{red}{$\downarrow$63.8\%}).
Code and models are available at \url{here}.

\end{abstract}

%% file: sec/1_intro.tex
\section{Introduction}
\begin{figure}[t]
	\centering
	\includegraphics[scale=0.43]{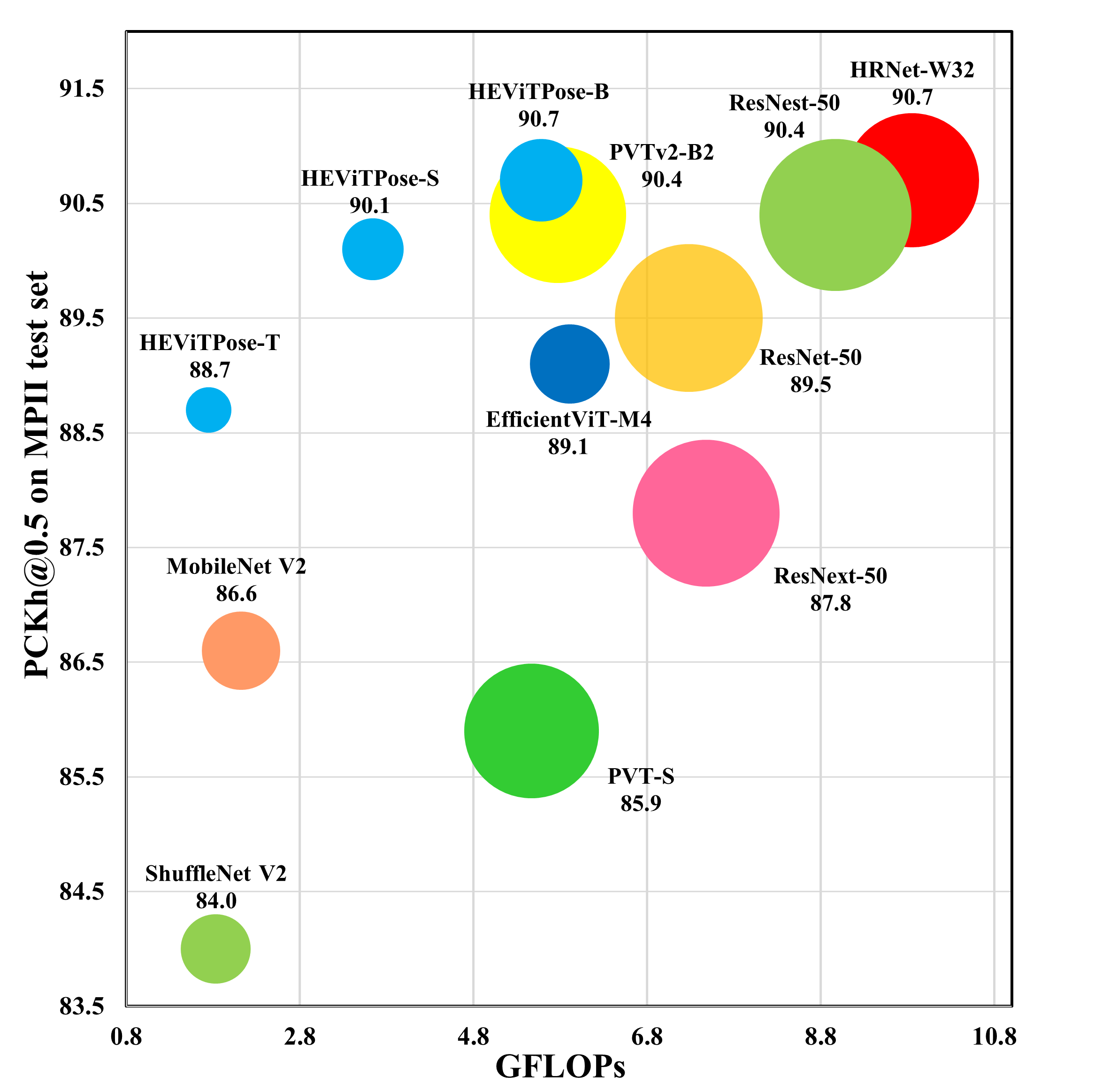}
	\caption{Comparison of HEViTPose and SOTA network models on the MPII test set regarding performance, parameters, and GFLOPs. The size of each bubble represents parameters.}
	\label{fig:PerformanceComparison}
\end{figure}

Human pose estimation (HPE) is a fundamental research topic in the field of computer vision and plays a crucial role in human-centred vision applications. The goal of HPE is to localize the exact pixel positions of keypoints of body parts from the images by detection and estimation methods. 
However, the development of HPE faces significant challenges when dealing with complicated situations, including viewpoint and appearance variations, occlusion, multiple persons, and imaging artefacts, etc.
With further research, HPE has a wide range of applications in 
action recognition \cite{duan2022revisiting}, 
intelligent surveillance \cite{khan2020human}, 
intention recognition \cite{fang2019intention}, 
and automated driving \cite{lu2020driver}, as one of the fundamental tasks of understanding human behaviour.

Developments in recent years have shown that deep learning-based methods have achieved state-of-the-art results in solving the Human pose estimation problem. There are currently two mainstream methods: (i) first predicting keypoints heatmap and then converting them to position \cite{xiao2018simple,newell2016stacked,chen2018cascaded,sun2019deep}, and (ii) directly regressing keypoints position \cite{toshev2014deeppose,sun2018integral,li2021human}.
In this paper, we study heatmap-based methods, which typically consist of a backbone network for feature extraction and a regressor for heatmap estimation.

According to the form of the different combinations of the backbone networks and the regressors, different types of network architectures are extended.
A common used network framework is to connect from high resolution to low resolution and then from low resolution to high resolution, such as SimpleBaseline \cite{xiao2018simple}, Hourglass \cite{newell2016stacked}.
Another network framework maintains high resolution throughout the process by connecting multi-resolution subnets in parallel, such as HRNet \cite{sun2019deep} and its variants \cite{yu2021lite,yuan2021hrformer}.
In addition, multi-scale fusion \cite{lin2017feature,chen2018cascaded,cheng2020higherhrnet} and multi-stage supervision \cite{wei2016convolutional} can be integrated into both types of network frameworks.
For intensive prediction tasks such as HPE, the ability of the backbone network to extract features often determines the performance of the model. Therefore, this paper adopts the SimpleBaseline \cite{xiao2018simple} configuration in its combinatorial form, using transposed convolution to accomplish the low-to-high process, and concentrates on the design of the backbone network.

The improvement of the performance of backbone networks mainly relies on the development of feature extraction techniques. 
For a long time, Convolutional Neural Network (CNN) have achieved remarkable success in computer vision due to their excellent feature extraction capability, becoming the dominant method in this field, such as \cite{he2016deep,newell2016stacked,ma2018shufflenet,sun2019deep,sandler2018mobilenetv2}.
However, the feature extraction capability of CNN is restricted by the receptive field. 
In order to extract long-range features, the receptive field has to be expanded by increasing the depth of the network, even if the feature contains only a small amount of information, which results in a larger network size and a higher computational overhead.
As a result, the CNN-based network model shows a significant increase in parameters and GFLOPs with gradual improvement in performance, such as MobileNetV2 \cite{sandler2018mobilenetv2}, ResNet-50 \cite{he2016deep}, and HRNet-W32 \cite{sun2019deep} in \cref{fig:PerformanceComparison}.
Recently, there has been a number of Transformer-based backbone networks proposed, which have received a lot of attention in the field of computer vision due to their excellent long-distance modelling capability and outstanding performance, such as 
\cite{dosovitskiy2020image,wang2021pyramid,han2021transformer,liu2021swin,liu2023efficientvit}.
The Transformer-based model outperforms the classical CNN-based model on large datasets. 
However, when the amount of data is insufficient, the Transformer-based model will fall behind due to the difficulty in exploiting its powerful feature extraction capability. 
As shown in \cref{fig:PerformanceComparison}, PVTv2-B2 \cite{wang2022pvt} slightly underperforms HRNet-W32 \cite{sun2019deep} of the same size in terms of performance on the MPII test set. It is worth noting that by adopting the Transformer-based approach we should take advantage of its long-range modelling capability rather than relying on a large number of block stacks.

In this study, we design the network architecture of HEViTPose for HPE tasks by taking inspiration from established networks (such as EfficientViT \cite{liu2023efficientvit}, Swin\cite{liu2021swin}, PVT \cite{wang2021pyramid}, and PVTv2 \cite{wang2022pvt}) to ensure a balance between model performance, size and computational overhead, as shown in \cref{fig:NetworkGraph2}. 
The HEViTPose shows a well-balanced performance in all aspects and surpasses all models shown in \cref{fig:PerformanceComparison}.

The main contributions of this work are summarized as:
\begin{itemize}[leftmargin=0.77cm, itemindent=0cm]
	\item We propose a CGSR-MHA module that combines the benefits of CGA \cite{liu2023efficientvit}, SRA \cite{wang2021pyramid}, and MHA \cite{vaswani2017attention}. This module significantly decreases computational costs by incorporating feature grouping and spatial degradation mechanisms, while maintaining feature diversity with multiple low-dimensional attention heads.
	\item We introduce the concept of PEOW based on OPE \cite{wang2022pvt}, further reveal the relationship between the number of overlapping edges and local continuity through experiments, and gradually improve the indicators of the model through the optimisation of PEOW.
\end{itemize}

\section{Related Work}
\textbf{Hunman Pose Estimation.}
The algorithmic frameworks for 2D multi-person pose estimation are classified into top-down \cite{sun2019deep,xiao2018simple,chen2018cascaded,fang2017rmpe} and button-up \cite{pishchulin2016deepcut,newell2017associative,geng2021bottom,luo2021rethinking}. 
The top-down algorithmic framework has been decomposing the multi-person pose estimation task into two sub-tasks, multi-person detection \cite{liu2016ssd,carion2020end,wang2023yolov7} and single-person pose estimation, which is considered to be high in accuracy, high in computation and slow in inference. While button-up algorithmic framework decomposes the task into two subtasks of keypoint detection and keypoint grouping for multiple people, which is considered to be computationally fast and less accurate.
In recent years, the continuous advancement of object detection algorithms \cite{arani2022comprehensive,lyu2022rtmdet,wang2020combining} has led to the promotion of the top-down algorithm framework, making a significant breakthrough in inference speed. As a result, it has gradually emerged in the task of real-time human pose estimation.
This paper conducts research on HPE via a top-down algorithmic framework, primarily concentrating on the architectural design of the backbone network.

\textbf{Transformer based vision backbones.}
The study of backbone architectures in this paper is an extension of ViT \cite{dosovitskiy2020image} and its related studies \cite{liu2021swin,touvron2021training,han2021transformer,wang2021pyramid}.
ViT \cite{dosovitskiy2020image} divides images into medium-sized image blocks and converts them into a series of fixed-length patch embeddings, and performs image classification through the Transformer architecture, achieving a balance between speed and accuracy.
However, the excellent performance of ViT relies heavily on the support of large-scale training datasets. 
In order to address this issue, DeiT \cite{touvron2021training} outlines various training approaches and distillation techniques that enhance data efficiency, thus rendering ViT more efficient when dealing with smaller datasets.
Several studies in the same period provided ideas to improve the performance of Transformer-based network architectures. 
PVT \cite{wang2021pyramid} introduces a pyramid structure to construct a multi-resolution feature map, which achieves better accuracy in dense prediction tasks.
LocalViT \cite{li2021localvit} incorporates depth-wise convolution into ViT to improve the local continuity of features.
Swin \cite{liu2021swin} adopts local window self-attention instead of global self-attention, reducing the quadratic relationship between network complexity and image size to a linear relationship and achieving a speed-accuracy balance.
In addition, MHSA \cite{vaswani2017attention} embeds the input features into multiple subspaces by attention head number and computes the attention maps separately, which has been shown to help improve model performance. However, improving performance simply by increasing the number of heads of attention is inefficient and creates significant computational redundancy.
The work of EfficientViT \cite{liu2023efficientvit} shows that assigning different splits of the complete feature to different attention heads can effectively reduce attentional computational redundancy.
This problem-solving approach follows the same line of thought as grouping convolution \cite{chollet2017xception,zhang2018shufflenet}.
In order to prevent the performance loss caused by excessive grouping, this work has employed two strategies. Firstly, it has appropriately increased the number of attention heads within the group. Secondly, it has controlled the dimensions of the Q, K, and V projections to correspond with the number of heads.
This approach can significantly reduce the computational overhead while ensuring the network performance.

\textbf{High resolution feature maps.}
The programme of high-resolution feature maps has been a great success on the HPE mission. In the development of high-resolution feature maps, four main approaches have emerged, including:
(i) Dilated convolutions \cite{yu2015multi,chen2018searching} maintain the high resolution of the feature map by removing some downsampling layers, preventing the loss of spatial information but incurring more computational cost.
(ii) Stacked Hourglass \cite{newell2016stacked}, CPN \cite{chen2018cascaded} utilise a decoder to restore high-resolution representations from low-resolution representations.
(iii) The high-resolution representation of the HRNet \cite{sun2019deep} model consists of different subnetworks with different resolutions, ensuring that the network retains its high resolution, and generates high-resolution feature maps with rich information through multi-scale fusion between branches.
(iv) Transposed convolution \cite{xiao2018simple,cheng2020higherhrnet} improves the resolution of the feature maps at the end of the network. SimpleBaselines \cite{xiao2018simple} demonstrates that transposed convolution can generate high-quality feature maps for heatmap prediction.
Our proposed HEViTPose follows the SimpleBaselines \cite{xiao2018simple} approach to generate high-resolution feature maps from low-resolution feature maps extracted from the backbone network by transposed convolution.

\section{Method}
\subsection{Overall Architecture}
\begin{figure}[!t]
	\centering
	\includegraphics[scale=0.16]{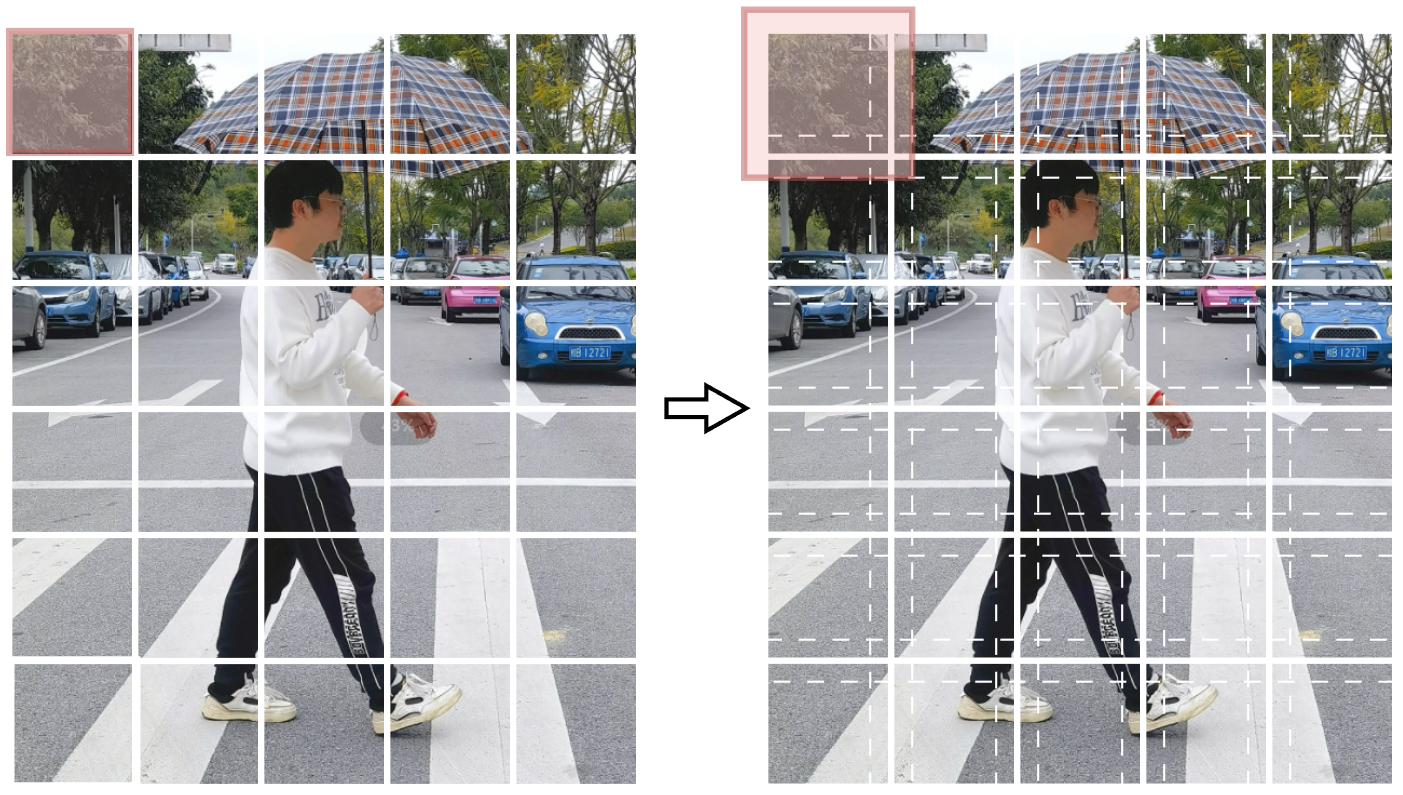}
	\caption{left: Patch Embedding of ViT; right: Overlapping Patch Embedding of PVTv2. The image block under the pink mask represents the first patch of each of these two methods.}
	\label{fig:people}
\end{figure}
\begin{figure*}[ht]
	\centering
	\includegraphics[scale=0.5]{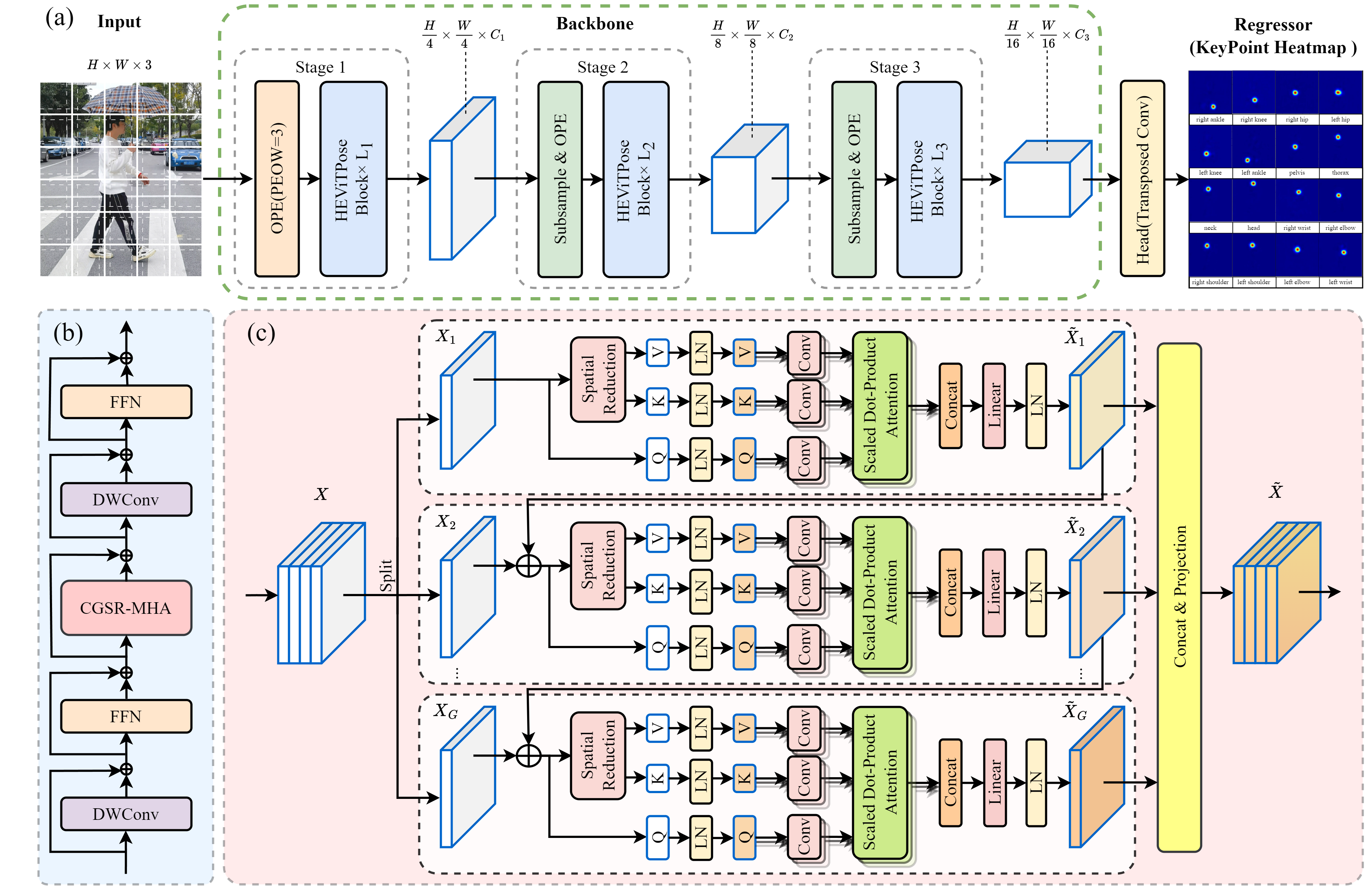}
	\caption{Overview of HEViTPose. (a) Network Architecture of HEViTPose; (b) HEViTPose Block; (c) CGSR-MHA.}
	\label{fig:NetworkGraph2}
\end{figure*}

For the HPE task, a network model called HEViTPose is designed in this paper, as shown in \cref{fig:NetworkGraph2}(a).
In the patch embedding part of the input image, we are inspired by the Overlapping Patch Embedding (OPE) \cite{wang2022pvt} to propose a concept of Patch Embedding Overlap Width (PEOW), and design an optimisation experiment of PEOW in \cref{sec:Experiments} to help readers further understand the relationship between the amount of overlap and local continuity. 
In the backbone network part, we implement the design concept of PVT \cite{wang2021pyramid}, which involves the incorporation of the progressive pyramid structure \cite{lin2017feature} into the Transformer framework. This enhances the performance of the HPE task by generating multi-scale feature maps. 
We have meticulously designed the Transformer-based backbone network also called HEViTPose. The network consists of three stages, each with a similar architecture, including a patch embedding layer, a downsampling layer (except for the first stage), and a transformer coding layer. When presented with an input image of size $H\times W\times 3$, the network generates three feature maps sequentially, producing a feature pyramid of $\frac{H}{4} \times \frac{W}{4} \times C_1$,  $\frac{H}{8} \times \frac{W}{8} \times C_2$, and  $\frac{H}{16} \times \frac{W}{16} \times C_3$.
In the head network part, we directly perform two up-sampling (transposed convolution) operations on the feature maps extracted from the backbone network, and the generated high-resolution feature maps ($\frac{H}{4} \times \frac{W}{4} \times 16$)\cite{xiao2018simple}.
In the regressor section, we simply regress the 16 keypoint heatmaps with $\frac{H}{4} \times \frac{W}{4} \times 16$ feature maps generated by the head network, while defining the loss function as the mean square error of the predicted heatmap and the groundtruth heatmap. Here the groundtruth heatmap is generated by a 2D Gaussian algorithm with a standard deviation of 1 pixel centred on the groundtruth position of each keypoint.

\subsection{Patch Embedding Overlap Width}
ViT \cite{dosovitskiy2020image} directly divides the image into non-overlapping patches, and then extracts embedding features for different patches separately through the Transformer network architecture, as shown in the left part of \cref{fig:people}. 
However, truncating the image causes a loss of continuity information between patches, making it challenging to reconstruct all continuity information even when combining various embedding features. 
PVTv2 \cite{wang2022pvt} preserves the local continuity of the image by extracting features through OPE, resulting in an improvement in the performance of the network, as shown in the right part of \cref{fig:people}.
Since the work of PVTv2 did not provide a detailed analysis of image patch overlap and local continuity, this section builds on OPE to portray the relationship between the amount of image patch overlap and local continuity through the relationship between the amount of image patch overlap and network performance.

\begin{figure}[!t]
	\flushleft
	\includegraphics[scale=0.29]{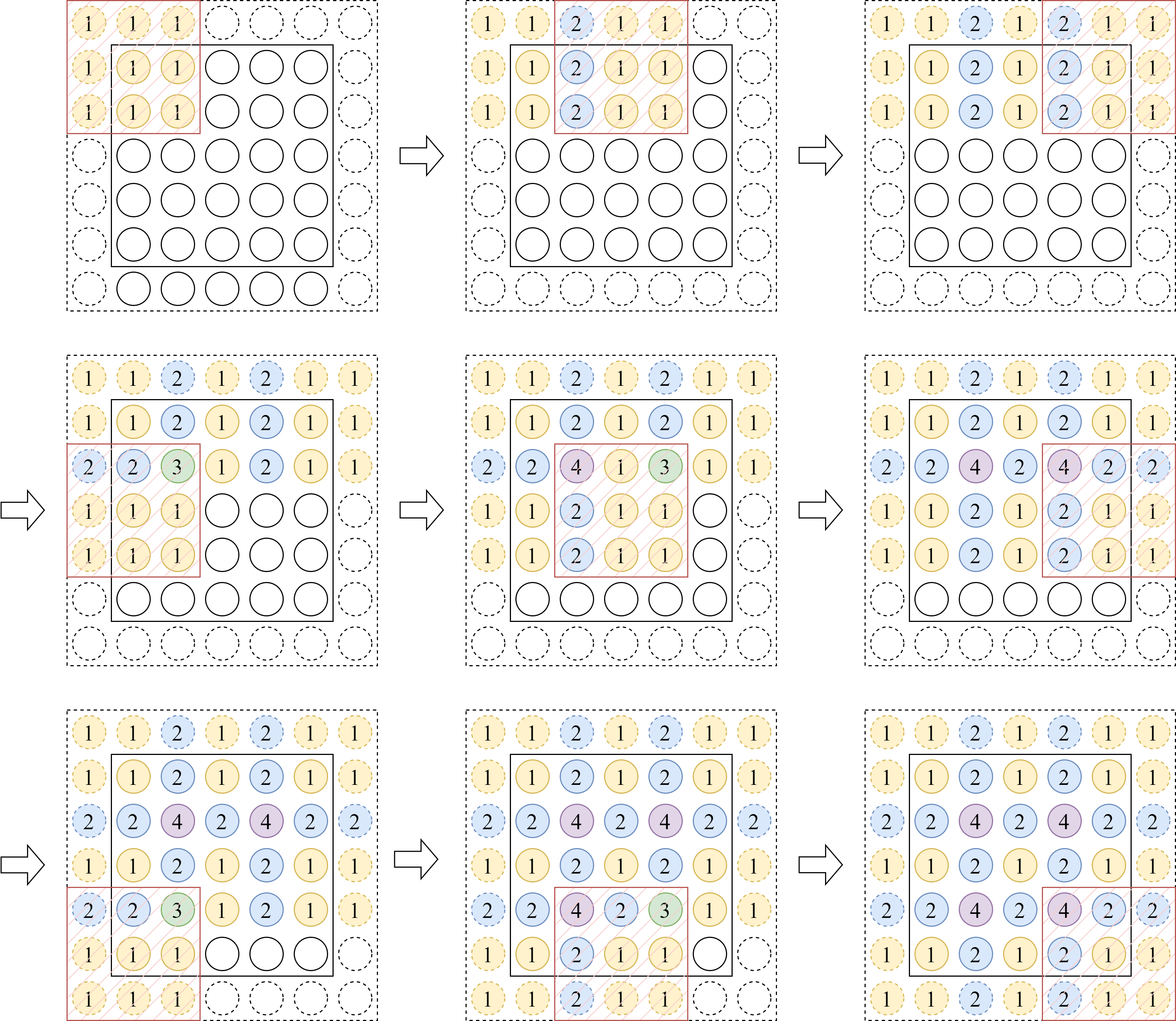}
	\caption{
		Set the convolution operation Conv2d(kernel\_size=3, stride=2, padding=1) corresponding to PEOW=1. Where solid circles indicate image pixels, dashed circles indicate padding pixels, and the 3$\times$3 mask indicates the convolution kernel.
	}
	\label{fig:analysis_1}
\end{figure}

\textbf{Definition of PEOW.}
In order to quantitatively describe the amount of overlap of image patches, the concept of patch embedding overlap width (PEOW) is proposed, i.e., the width of the "grid lines" formed by repeated computations of the convolution kernel and image pixels. \cref{fig:analysis_1} shows the case of PEOW = 1, where the number in the circle indicates the number of times an image pixel is used for computation when it is convolved.

\begin{figure}[!t]
	\centering
	\includegraphics[scale=0.19]{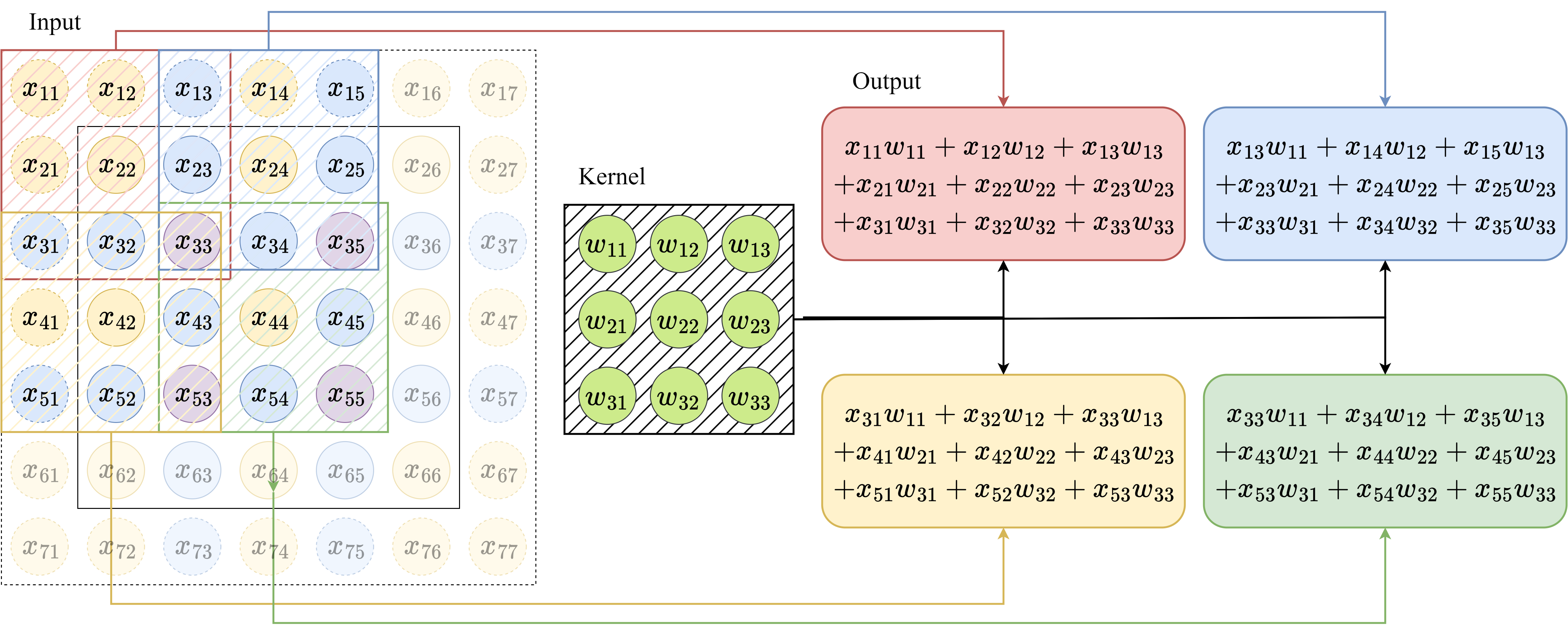}
	\caption{Modelling process for convolution with PEOW=1. For ease of understanding, here let the bias b = 0.}
	\label{fig:analysis_2}
\end{figure}

\textbf{Analysis.}
By observing the sliding process of the convolution kernel in \cref{fig:analysis_1}.
We find that pixels with the number 1 are associated with elements in only 1 region covered by the convolution kernel, pixels with the number 2 can be associated with elements in 2 regions, and pixels with the number 4 can be associated with 4 regions.
The modeling process of a convolution kernel is shown in \cref{fig:analysis_2}. The continuity between two adjacent image patches can be modeled through three pixel point samples. The continuity between four pairwise adjacent image patches can only be modeled by a sample of 1 pixel point.

Assuming that the modelling function $\mathcal{F}(\cdot)$ in this layer contains operations such as convolution, activation function $act(\cdot)$, etc., the predicted values of the four outputs in \cref{fig:analysis_2} are $\hat y_1$, $\hat y_2$, $\hat y_4$, $\hat y_5$, as in \cref{eq:pred-1,eq:pred-2,eq:pred-3,eq:pred-4}.
Assuming that the overall modelling function is $\mathcal{G}(\cdot)$, the final output of any of the predicted values is $\hat z_i$, as in \cref{eq:pred-5}.
\begin{align}
	\hat y_1 &= act(x_{11}w_{11}+...+x_{33} w_{33} ) \notag \\ 
	         &= \mathcal{F}(x_{13}, x_{23}, x_{31}, x_{32}, x_{33}) 
	\label{eq:pred-1}	\\
	\hat y_2 &= \mathcal{F}(x_{13}, x_{23}, x_{33}, x_{34}, x_{35})
	\label{eq:pred-2} 	\\
	\hat y_4 &= \mathcal{F}(x_{31}, x_{32}, x_{33}, x_{43}, x_{53})
	\label{eq:pred-3} 	\\
	\hat y_5 &= \mathcal{F}(x_{33}, x_{34}, x_{35}, x_{43}, x_{53}) 
	\label{eq:pred-4}	\\
	\hat z_i &= \mathcal{G}(\hat y_1,\hat y_2,…,\hat y_4,\hat y_5,…)
	\label{eq:pred-5}	
\end{align}

We can observe from \cref{eq:pred-1,eq:pred-2,eq:pred-3,eq:pred-4,eq:pred-5} that although the deep model $\mathcal{G}$ has all the input variables, it is relatively difficult to establish continuity through different layers of variables nested in functions. However, we can provide a wealth of information for building local continuity by controlling PEOW in the shallow model $\mathcal{F}$. Thus, two points for further consideration arise.
(i) Although PEOW=1 can provide local continuity information, the shallow model is limited in its ability to capture continuity due to the small amount of information provided.
(ii) For models supervised by Gaussian heatmaps, the derivative information around the mean is also an important part of what constitutes an accurate prediction, which cannot be modelled by PEOW=1.

\begin{figure}[!t]
	\centering
	\subfloat[]{
		\label{analysis_3}
		\includegraphics[width=0.25\linewidth]{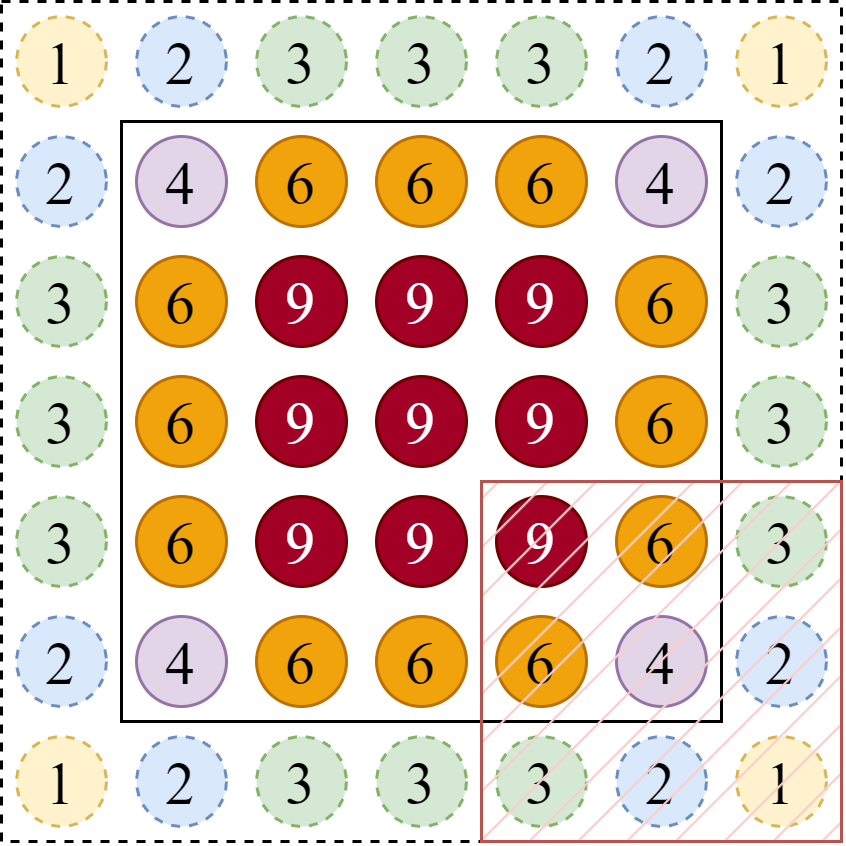}
	}
	\subfloat[]{
		\label{analysis_4}
		\includegraphics[width=0.535\linewidth]{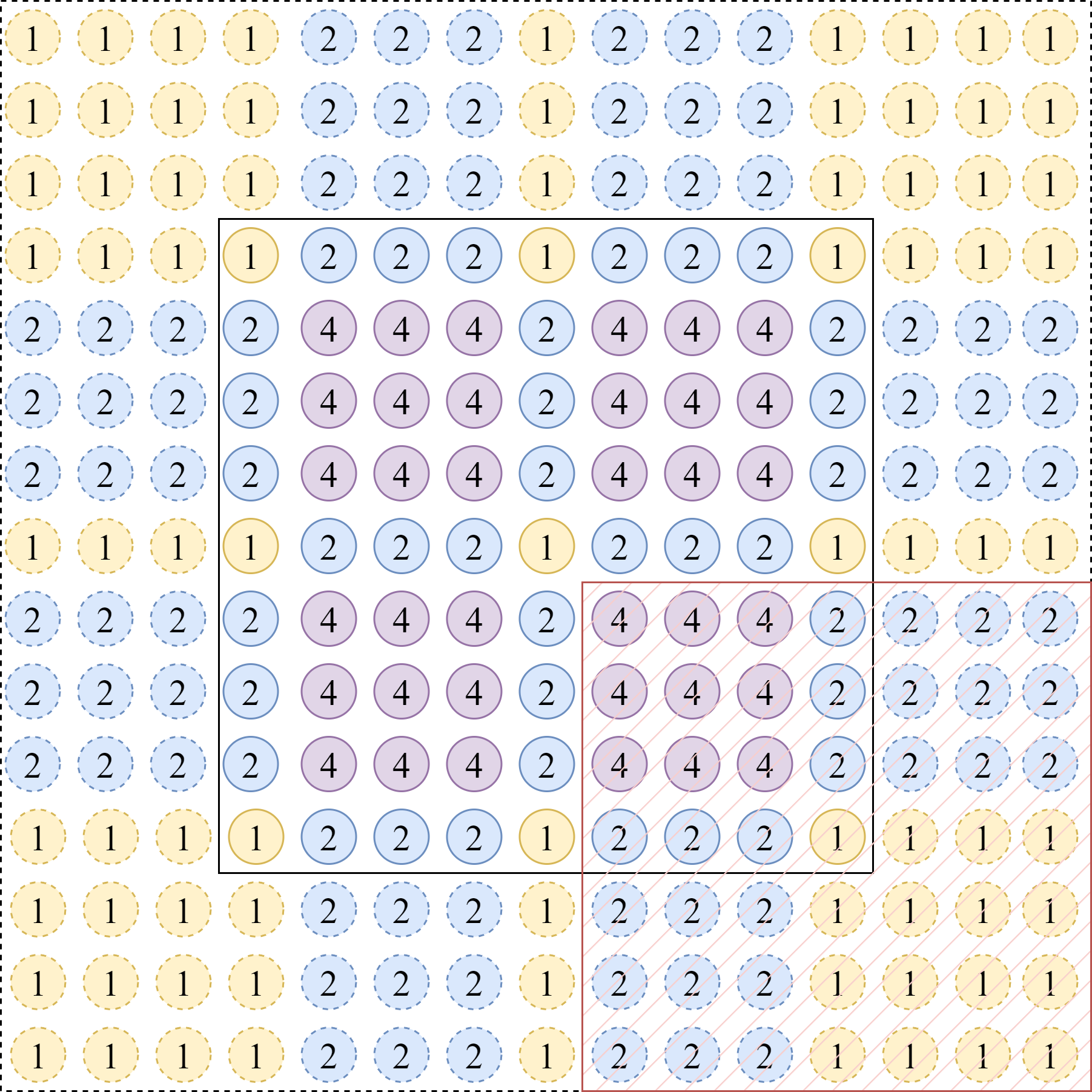}
	} 	
	\caption{Two ways to add PEOW. 
		(a) Setting the multilayer overlap corresponding to the convolution operation Conv2d(3,1,1);
		(b) Setting PEOW=3 corresponding to the convolution operation Conv2d(7,4,3).}
	\label{fig:analysis}
\end{figure}

It is not challenging for us to suggest expanding PEOW to improve continuity and conductivity information. We did not analyze the stride value reduction method for increasing PEOW, because this method introduces multilayer overlapping, which easily produces information redundancy, as shown in \cref{fig:analysis} (a). Therefore, we adopt the method of increasing the size of the convolution kernel to increase PEOW, as shown in \cref{fig:analysis} (b). In the experimental section, we select different PEOWs to conduct experiments to verify the reasonableness of the thinking.

\subsection{HEViTPose Building Block}
In order to achieve a balance between model performance and feature extraction efficiency, we first introduce a sandwich layout from EfficientViT \cite{liu2023efficientvit} in the HEViTPose building blocks to reduce the memory time consumption caused by the self-attentive layer in the model and to enhance communication between channels.
Secondly, we follow the PVTv2 \cite{wang2022pvt} approach by replacing the fixed-size positional embedding with the positional coding introduced by the zero-padded convolutional layer (DWConv) to adapt to arbitrary size input images, as shown in \cref{fig:NetworkGraph2} (b).
Thirdly, we have utilised the successful practices of 
TNT \cite{han2021transformer}, MHSA \cite{vaswani2017attention},  CGA \cite{liu2023efficientvit}, ISSA \cite{huang2019interlaced}, SRA \cite{wang2021pyramid}, 
and other attentional mechanisms to propose a Cascaded Group Spatial Reduction Multiple Head Attention (CGSR-MHA), which is more accurate in terms of precision, more computationally efficient, and more suitable for such intensive tasks as human pose estimation, by providing each subgroup with a different segmentation of the full feature and thus explicitly decomposing the attentional computation between subgroups, as in \cref{fig:NetworkGraph2}(c).

\textbf{CGSR-MHA.}
\begin{table*}[!t]
	\centering
	\caption{\textbf{Comparison on the MPII test set(PCKh@0.5).}  The performance, parameters, and GFLOPs for the pose estimation network are measured w/o considering human detection. All results come from retraining under the same conditions, and none uses any pre-training. We compute the percentages in terms of parameters and GFLOPs reduction between models marked with the same symbol.}
	\setlength{\tabcolsep}{1.6mm}{
		\begin{tabular}{l|c|c|c c c c c c c|c}
			\toprule[1.5pt]
			Method 									 	&Params	&FLOPs 	&Hea	&Sho 	&Elb 	&Wri 	&Hip 	&Kne 	&Ank 	&Total\\
			\midrule[1.5pt]
			ShuffleNetV2\cite{ma2018shufflenet}		&7.55M	&1.83G	&96.6	&92.6	&83.8	&76.6	&84.9	&77.6	&71.4	&84.0\\
			MobileNetV2\cite{sandler2018mobilenetv2}	&9.57M\S	&2.12G\S	&97.3	&94.5	&86.8	&80.7	&87.3	&80.9	&74.6	&86.6\S\\		
			ResNest-50\cite{zhang2022resnest}			&35.93M	&8.97G	&97.9	&96.3	&90.9	&86.0	&89.8	&86.7	&82.1	&90.4\\
			ResNet-50\cite{he2016deep}					&34.0M	&7.28G	&97.7	&95.8	&89.8	&84.5	&89.8	&85.7	&79.9	&89.5\\
			ResNext-50\cite{xie2017aggregated}			&33.47M	&7.48G	&97.9	&95.8	&89.8	&84.4	&89.5	&85.8	&80.4	&89.5\\
			HRNet-W32\cite{sun2019deep}					&28.02M\dag &9.85G\dag	&\textbf{98.2}	&\textbf{96.2}	&\textbf{91.3}	&\textbf{87.1}	&90.1	&\textbf{86.8}	&82.4	&\textbf{90.7}\\		
			PVT-S\cite{wang2021pyramid}					&28.17M	&5.47G	&96.9	&93.6	&86.0	&79.8	&86.2	&79.8	&74.4	&85.9\\
			Swin-S\cite{liu2021swin}			&54.1M	&15.4G	&97.5	&94.8	&88.4	&83.0	&88.4	&83.5	&78.4	&88.2\\
			EfficientViT-M4\cite{liu2023efficientvit}	&9.87M	&5.91G	&97.6	&95.2	&89.4	&84.3	&89.1	&84.5	&80.3	&89.1\\
			PVTv2-B2\cite{wang2022pvt}					&29.05M\ddag	&5.77G\ddag	&98.0	&96.3	&90.7	&85.9	&90.2	&86.3	&82.4	&90.4\\
			\midrule	
			\textbf{HEViTPose-T}						&3.21M\S(\textcolor{red}{$\downarrow$66.5\%})	&1.75G\S(\textcolor{red}{$\downarrow$17.5\%})	&97.6	&95.1	&89.0	&83.6	&89.1	&83.9	&79.1	&88.7\S(\textcolor{red}{$\uparrow$2.1})\\
			\textbf{HEViTPose-S}						&5.88M\ddag(\textcolor{red}{$\downarrow$79.8\%})	&3.64G\ddag(\textcolor{red}{$\downarrow$37.0\%})	&97.8	&95.9	&90.5	&86.0	&89.7	&86.0	&81.7	&90.1\\
			\textbf{HEViTPose-B}						&\textbf{10.63M}\dag(\textcolor{red}{$\downarrow$62.1\%})	&\textbf{5.58G}\dag(\textcolor{red}{$\downarrow$43.4\%})	&98.0	&96.1	&\textbf{91.3}	&86.5	&\textbf{90.2}	&86.6	&\textbf{83.0}	&\textbf{90.7}\\	
			\bottomrule[1.5pt]
		\end{tabular}
	}
	\label{tab:mpii_test}
\end{table*}
The Cascaded Group Attention (CGA) proposed by EfficientViT \cite{liu2023efficientvit} alleviates the attention head redundancy problem of Multiple Head Self-Attention (MHSA), resulting in an improvement in feature extraction efficiency. However, the CGA method exclusively divides each cascade group feature differently into Q, K, and V matrices for ablation experiments, which is more conservative in solving the information redundancy problem. Therefore, this paper adopts the idea of PVT \cite{wang2021pyramid} to reduce the dimensionality of the features within the cascade group, and solves the information redundancy problem by directly controlling the dimensionality of the Q, K, V matrices. 
Formally, CGSR-MHA can be formulated as:

We divide the input feature map $X\in\mathbb{R}^{C\times H \times W}$ into $G$ groups: $X=[X_1,X_2,...,X_G]$, where $X_g\in\mathbb{R}^{\frac{C}{G}\times H\times W}$ denotes the feature map for each cascade grouping and $g\in[1,2,...,G]$ denotes the number of each cascade grouping.
\begin{gather}
	\tilde{X}_{g} =\left \{
	\begin{aligned}
		\mathcal{L}(X_g, r) , g & =  1 \\
		\mathcal{L}(X_{g}+\tilde{X}_{g-1}, r) , g & \in \left ( 2,3,\cdots ,G \right )  
	\end{aligned}
	\right.
	\label{eq:pred-6}		
\end{gather}
where $\mathcal{L}(\cdot)$ denotes the attention function within the cascade group, and $X_g \in\mathbb{R}^{\frac{C}{G}\times H\times W}$  denotes the feature map within the cascade group after $\mathcal{L}(\cdot)$ processing.
\begin{equation}
	\mathcal{L}(X_g, r) = Rep_{l}^{hw}(MHA( Rep_{hw}^{l}( SR( X_g, r) ) ) ),
	\label{eq:eq7}	
\end{equation}
where $SR(\cdot)$ denotes the spatial reduction operation and $r$ denotes the spatial reduction rate. 
$Rep_{hw}^l(\cdot)$ denotes the operation of reshaping the features in the group of size $\frac{C}{G} \times H \times W$ to $\frac{C}{G} \times L$ first, and then the $LN(\cdot)$ operation is performed on the feature map, where $L= H\times W$ denotes the encoding length and $LN(\cdot)$ denotes the layer normalisation operation.
$Rep_l^{hw}(\cdot)$ denotes the operation of performing $LN(\cdot)$ on the feature map first, and then the operation of reshaping the features of size $\frac{C}{G} \times L$ to  $\frac{C}{G} \times H \times W$.

\begin{equation}
	\tilde{X} = Linear(Concat(\tilde{X}_{1},\tilde{X}_{2},...,\tilde{X}_{G})),
	\label{eq:eq8}	
\end{equation}
where $Linear(\cdot)$ denotes the linear projection operation and $Concat(\cdot)$ denotes the concatenation operation. $\tilde{X}$ denotes the output feature map, obtained by the CGSR-MHA($\cdot$) operation.

\textbf{Model families of HEViTPose.}
In order to compare the network models of different scale, we build a model family containing three models and show the details of the architecture of each model in \cref{tab:model_family}. $C_i$, $L_i$, $G_i$, $R_i$ are the width, depth, number of groupings and spatial reduction rate of the ith stage, respectively.
\begin{table}[!t]
	\centering
	\caption{Architectural details of the HEViTPose model family.}
	\setlength{\tabcolsep}{0.1mm}{
		\begin{tabular}{c|c|c|c}
			\toprule[1.5pt]
			Model					&HEViTPose-T		&HEViTPose-S		&HEViTPose-B	\\
			\midrule[1.5pt]
			$\{C_1,C_2,C_3\}$		&\{64,128,192\}		&\{128,192,224\}	&\{128,256,384\}\\					
			$\{G_1,G_2,G_3\}$		&\{1,2,3\}			&\{1,2,3\}			&\{1,2,3\}		\\
			$\{L_1,L_2,L_3\}$		&\{4,4,4\}			&\{4,3,2\}			&\{4,4,4\}		\\					
			$\{R_1,R_2,R_3\}$		&\{8,4,2\}			&\{8,4,2\}			&\{8,4,2\}		\\							
			\bottomrule[1.5pt]
		\end{tabular}
	}
	\label{tab:model_family}
\end{table}

\section{Experiments}
\label{sec:Experiments}
\subsection{MPII Human Pose Estimation}
\textbf{Dataset.}
The MPII Human Pose dataset \cite{mmpose2020} has around 28k person instances used as training samples and around 12k person instances used as test samples.
\begin{table*}[ht]
	\centering
	\caption{\textbf{Comparison on the COCO test-dev2017 set.} The performance, parameters, and GFLOPs for the pose estimation network are measured w/o considering human detection. All results come from retraining under the same conditions, and none uses any pre-training. We compute the percentages in terms of parameters and GFLOPs reduction between models marked with the same symbol.}
	\setlength{\tabcolsep}{3.5mm}{
		\begin{tabular}{l|c|c|c c c c c c}
			\toprule[1.5pt]
			Method 										&Params		&FLOPs 	&AP		&AP\textsuperscript{50}	&AP\textsuperscript{75}	&AP\textsuperscript{M}	&AP\textsuperscript{L}	&AR\\
			\midrule[1.5pt]
			ResNet-50\cite{he2016deep}					&34.0M		&7.28G	&69.6	&91.0	&77.6	&66.2	&75.5	&75.3\\	
			PVT-S\cite{wang2021pyramid}					&28.17M	&5.47G	&68.9	&90.9	&77.4	&65.7	&74.7	&74.8\\
			ViTPose-B\cite{xu2022vitpose}				&90.04M	&23.8G	&70.9	&91.3	&79.6	&68.4	&76.0	&76.8\\
			Swin-S\cite{liu2021swin}			&54.1M\dag	&15.4G\dag	&\textbf{72.7}	&92.1	&81.5	&69.5	&78.4	&78.3\\
			PVTv2-B2\cite{wang2022pvt}					&29.05M	&5.77G	&\textbf{72.7}	&92.1	&81.4	&69.3	&78.4	&78.1\\
			\midrule
			\textbf{HEViTPose-B}						&\textbf{10.63M}\dag(\textcolor{red}{$\downarrow$80.4\%})	&\textbf{5.58G}\dag(\textcolor{red}{$\downarrow$63.8\%})	&\textbf{72.6}	&92.0	&80.9	&69.2	&78.2	&78.0\\	
			\bottomrule[1.5pt]
		\end{tabular}
	}
	\label{tab:coco_test}
\end{table*}

\textbf{Training details.}
In this paper, we follow the common training strategy of the mmpose codebase \cite{mmpose2020}, setting up different data pipelines for the training set and the validation set.
The model was trained on NVIDIA RTX3060 GPU (12GB). We use the Adam \cite{kingma2014adam} optimizer. The learning schedule follows the setting \cite{papandreou2017towards}. The base learning rate is set to 1e-3, and dropped to 1e-4 and 1e-5 at the 170th and 200th epochs, respectively. The training process is terminated within 210 epochs. 
We set the input size to 256×256 and the training batch size to 32.
Note that all models are trained from scratch without any pre-training.

\textbf{Testing details.}
We follow the two-stage top-down multiple human pose estimation paradigm similar as \cite{papandreou2017towards,chen2018cascaded}, which consists of using object detectors to detect human instances and using pose estimation networks to generate keypoint predictions for the instances.
We use the same person detectors provided by SimpleBaseline \cite{xiao2018simple} for both the validation set and the test set. Following \cite{newell2016stacked,chen2018cascaded}, we first compute the heatmap by averaging the heatmap of the original image and the flipped image, and then convert the predicted heatmap into position coordinates using the standard decoding method. 

\textbf{ Results on the test set.}
\cref{tab:mpii_test} reports the human pose estimation performance of our method and existing state-of-the-art methods on the MPII test set. Our proposed HEViTPose family achieves competitive performance compared with the state-of-the-art methods in terms of fewer model size (Params) and computation complexity (GFLOPs).
For example, our HEViTPose-B achieves a score of 90.7 PCKh@0.5, which is the same as HRNet-W32 \cite{sun2019deep} with 62.1\% fewer parameters and 43.4\% fewer GFLOPs.
Compared to PVTv2 \cite{wang2022pvt}, our HEViTPose-S reduces only 0.3 AP, but results in 79.8\% fewer parameters and 37.0\% fewer GFLOPs.
Compared to MobileNetV2 \cite{sandler2018mobilenetv2}, our HEViTPose-T improves by 2.1 AP with 66.5\% fewer parameters and 17.5\% fewer GFLOPs.

\subsection{COCO Keypoints Detection}
\textbf{Dataset.}
The COCO dataset \cite{lin2014microsoft} was divided into train2017, val2017 and test-dev2017 sets with 57k, 5k and 20k images, respectively.

\textbf{Training and testing details.}
In training and testing, we set the data enhancement pipeline, training strategy, and human pose estimation paradigm the same as MPII. But in order to speed up the training process of the model on the COCO 2017 dataset, we set the batch size to 64 and distributed the model training on 2 NVIDIA RTX4090 GPUs (2×24GB). 

\textbf{Results on the test-dev2017 set.}
As shown in \cref{tab:coco_test}, we compared our HEViTPose-B with several of the most representative networks on COCO test-dev2017, and our network showed competitive results on parameters and GLOPs. For example,
compared to the excellent Swin-S \cite{liu2021swin}, our HEViTPose-B is only 0.1 AP lower, while the parameters are 80.4\% lower and the GFLOPs are 63.8\% lower.
Compared with the most competitive PVTv2-B2 \cite{wang2022pvt} in recent times, our HEViTPose-B is only 0.1 AP lower, while the parameters are 63.4\% lower and the GFLOPs are 3.3\% lower.

\begin{table}[!t]
	\centering
	\caption{Influence of adjusting the PEOW value on the HEViTPose network on the MPII validation set(PCKh@0.5). The input image size is 3×256×256 and the output image size is 128×64×64.}
	\setlength{\tabcolsep}{1mm}{
		\begin{tabular}{c|c|c|c|c}
			\toprule[1.5pt]
			PEOW	&(kernel\_size, stride)	&Params	&FLOPs		&Total\\
			\midrule[1.5pt]	
			1		&conv(3,2), conv(3, 2)		&9.87M					&5.91G				&87.6\\
			\midrule
			3		&conv(7, 4)					&\textbf{9.82}			&\textbf{5.65G}		&\textbf{88.1}\\
			\midrule
			7		&conv(15,8), deconv(4,2)	&10.5M					&6.29G				&85.0\\
			\midrule
			3		&conv(7, 2), conv (7, 2)	&10.21M					&7.38G				&87.5\\
			\bottomrule[1.5pt]
		\end{tabular}
	}
	\label{tab:PEOW}
\end{table}
\begin{table*}[!t]
	\centering
	\caption{Results of different group numbers in each stage of HEViTPose on MPII validation set(PCKh@0.5).
		$\{g_1,g_2,g_3\}$: the number of cascade groups in each stages; 
		$\{r_1,r_2,r_3\}$: the spatial reduction ratio of the features within the group at each stage; 
		$\{h_1,h_2,h_3\}$: the number of attention heads for spatial reduction features within the group for each stage.}
	\setlength{\tabcolsep}{3mm}{
		\begin{tabular}{c|c|c|c|c|c|c|c}
			\toprule[1.5pt]			
			\# 						&$\{g_1,g_2,g_3\}$ &$\{r_1,r_2,r_3\}$ &$\{h_1,h_2,h_3\}$ &Params 	&FLOPs 	&Training Time(day)	&Total(PCKh@0.5)	\\	
			\midrule[1.5pt]	
			\multirow{3}{*}{1}  	&\{2,2,2\}	&\{8,4,2\}  &\{2,4,8\}    &13.72M 						&5.80G 						&1							&89.1	\\
			\cline{2-8}				&\{4,4,4\}	&\{8,4,2\}	&\{2,4,8\}	&10.63M						&5.58G						&\textbf{1}					&\textbf{89.4}	\\
			\cline{2-8} 			&\{8,8,8\}	&\{8,4,2\} 	&\{2,4,8\} 	&\textbf{9.86M} 				&\textbf{5.46G} 				&3							&89.0	\\	
			\midrule
			\multirow{3}{*}{2}  	&\{4,4,4\}	&\{4,2,1\}	&\{2,4,8\}	&9.87M						&5.48G						&10							&-				\\
			\cline{2-8}				&\{4,4,4\}	&\{8,4,2\}	&\{2,4,8\}	&10.63M						&5.58G						&\textbf{1}					&\textbf{89.4}	\\
			\cline{2-8} 			&\{4,4,4\}	&\{16,8,4\}	&\{2,4,8\}	&13.94M						&5.62G						&1							&89.0			\\	
			\midrule
			\multirow{2}{*}{3}  	&\{4,4,4\}	&\{4,2,1\}	&\{1,2,4\}	&9.87M						&5.48G						&3							&89.3	\\
			\cline{2-8}				&\{4,4,4\}	&\{4,2,1\}	&\{2,4,8\}	&9.87M						&5.48G						&10							&-		\\
			\bottomrule[1.5pt]
		\end{tabular}
	}
	\label{tab:HEViT_1}
\end{table*}

\subsection{Ablation Experiments}
In this subsection, we investigate the effect of each component in HEViTPose on the MPII human pose estimation dataset. All settings follow the MPII experiments.

\textbf{Influence of PEOW.}
In \cref{tab:PEOW}, we investigate the effect of different PEOW on HEViTPose. We observe that proper control of PEOW can significantly improve the model performance while reducing parameters and GFLOPs.
According to \cref{tab:PEOW}, we have the following findings:
(i) We can obtain the highest PCKh@0.5 score of 88.1 when we control PEOW = 3 and when stride = 4, while obtaining the lowest parameters (9.82M) and the lowest GFLOPs (5.65G).
(ii) For double-layer overlap, the PCKh@0.5 score increases by 0.5 when PEOW is increased from 1 to 3. When PEOW is increased from 3 to 7, the PCKh@0.5 score decreases by 3.1. This suggests that there is a limit to improving the network performance by increasing PEOW alone, and that choosing the right PEOW will allow the model to maintain a balanced performance, parameters, and GFLOPs.
(iii) Controlling PEOW to be 3, comparing stride to be 2 and 4, we find that the multi-layer overlap created by stride to be 2 does not improve the PCKh@0.5 score, but rather reduces it by 0.6, and greatly increases parameters and GFLOPs.

\textbf{Influence of HEViTPose.}
In \cref{tab:HEViT_1}, we adjust the parameter configurations of the HEViTPose model and observe the variation of the model over the MPII val set.
As described in \#1, when $\{r_1,r_2,r_3\}$ and $\{h_1,h_2,h_3\}$ are unchanged, the larger $\{g_1,g_2,g_3\}$ is, the lower the parameters and GFLOPs of HEViTPose-B.
As described in \#2, when $\{g_1,g_2,g_3\}$ and $\{h_1,h_2,h_3\}$ are unchanged, the larger $\{r_1,r_2,r_3\}$ is, the larger the parameters and GFLOPs of HEViTPose-B are, but the training time of the model is also drastically reduced.
As described in \#3, when $\{g_1,g_2,g_3\}$ and $\{r_1,r_2,r_3\}$ are constant, the training time increases substantially as $\{h_1,h_2,h_3\}$ increases. Due to limited training resources, we adjust the number of feature heads in the group to keep normal operation.
After experiments, we obtain a set of parameter values that are balanced in performance, parameters, GFLOPs and training efficiency: $\{g_1,g_2,g_3\}= \{4,4,4\}$, $\{r_1,r_2,r_3\}= \{8,4,2\}$, $\{h_1,h_2,h_3\}= \{2,4,8\}$. At this point EViTPose-B has the best performance at 89.4 PCKh@0.5 while keeping low parameters, GFLOPs and training time.

\textbf{Ablation of HEViTPose on MPII validation set.}
In \cref{tab:Ablation}, we compare the optimization effects of different components on the model on the MPII validation set. Finally, under the condition that the number of parameters is maintained, the PCKh@0.5 score is increased by 1.8, and the computation amount is decreased by 5.6\%. 
The implementation details of the different methods in the table are as follows:
(a) The patch embedding part follows the OPE configuration of EfficientViT \cite{liu2023efficientvit}, which corresponds to the case of PEOW of 1 as proposed in this paper, with EfficientViT-M4 for the backbone network, and other configurations follow the basic Top-Down paradigm adopted. 
(b) Adjusting PEOW=3 enhances the PCKh@0.5 score of this network model by 0.5 to 88.1. It also lowers GFLOPs by 4\% and slightly reduces the number of parameters.
(c) The backbone network replaces EfficientViT-M4 with the HEViTPose-B proposed in this paper, which further improves the model PCKh@0.5 score by 1.3 and decreases GFLOPs by 1.2\%.
\begin{table}[!t]
	\centering
	\caption{Ablation experiments for HEViTPose-B on the MPII validation set(PCKh@0.5).}
	\setlength{\tabcolsep}{0.9mm}{
		\begin{tabular}{c|c|c|c|c|c}
			\toprule[1.5pt]
			Method	&PEOW=3		&HEViTPose	&Params						&FLOPs 						&Total\\
			\midrule[1.5pt]		
			(a)		&\ding{55}	&\ding{55}	&9.87M 						&5.91G						&87.6\\
			\midrule		
			(b)		&\ding{51}	&\ding{55}	&9.82M						&5.65G						&88.1\\
			\midrule		
			(c)		&\ding{51}	&\ding{51}	&10.63M						&\textbf{5.58G}				&\textbf{89.4}\\
			\bottomrule[1.5pt]
		\end{tabular}
	}
	\label{tab:Ablation}
\end{table}

\section{Conclusion}
The paper first presents the concept of Patch Embedding Overlap Width (PEOW), which can help readers to further understand the role of Overlapping Patch Embedding (OPE) and provides an effective tool for adjusting the amount of overlap to re-establish local continuity.
Then, the text proposes the High-Efficiency Vision Transformer for Human pose estimation (HEViTPose), which is a high-performance and efficient transformer architecture. The key idea is to reduce the computational redundancy through feature grouping and in-group feature dimensionality reduction, while retaining high performance through cascading of grouping and MHA of in-group features, which improves the efficiency of feature extraction.
Finally, our HEViTPose benefits from the information provided by the early convolution containing local continuum features, and also benefits from the remote information interaction of the transformer in the cascade group. 
We experimentally validate the effectiveness of HEViTPose on a pose estimation task.

\section*{Acknowledgement}
This work was partially supported by National Natural Science Foundation of China under grants 61563005.